\def\year{2022}\relax
\documentclass[letterpaper]{article} 
\usepackage{aaai22}  
\usepackage{times}  
\usepackage{helvet}  
\usepackage{courier}  
\usepackage[hyphens]{url}  
\usepackage{graphicx} 
\usepackage{natbib}  
\usepackage{caption} 
\DeclareCaptionStyle{ruled}{labelfont=normalfont,labelsep=colon,strut=off} 
\usepackage{algorithm}
\usepackage{algorithmic}

\usepackage{newfloat}
\usepackage{listings}
\floatstyle{ruled}
\newfloat{listing}{tb}{lst}{}
\floatname{listing}{Listing}
\title{Improving Behavioural Cloning with Human-Driven Dynamic Dataset Augmentation}
\author {
    Federico Malato\textsuperscript{\rm 1}\thanks{Accepted at the AAAI-22 Workshop on Interactive Machine Learning (IML@AAAI’22)},
    Joona Jehkonen\textsuperscript{\rm 1},
    Ville Hautamäki\textsuperscript{\rm 1,2}
}
\affiliations {
    \textsuperscript{\rm 1} School of Computing, University of Eastern Finland\\
    \textsuperscript{\rm 2}Department of Electrical and Computer Engineering, National University of Singapore\\
    federico.malato@uef.fi, joojeh@student.uef.fi, ville.hautamaki@uef.fi
}

\usepackage{bibentry}

\begin{document}

\maketitle

\begin{abstract}
Behavioural cloning has been extensively used to train agents and is recognized as a fast and solid approach to teach general behaviours based on expert trajectories. Such method follows the supervised learning paradigm and it strongly depends on the distribution of the data. In our paper, we show how combining behavioural cloning with human-in-the-loop training solves some of its flaws and provides an agent task-specific corrections to overcome tricky situations while speeding up the training time and lowering the required resources. To do this, we introduce a novel approach that allows an expert to take control of the agent at any moment during a simulation and provide optimal solutions to its problematic situations. Our experiments show that this approach leads to better policies both in terms of quantitative evaluation and in human-likeliness.
\end{abstract}

\noindent In {\em reinforcement learning} (RL) \cite{SuttonBarto1998} the difficulty for training an agent that can successfully solve a task is directly proportional to the complexity of the task itself. This happens because RL is based on the concept of reward function, that is, a signal that should encourage a correct behaviour while discouraging a harmful one at the same time. In a standard RL task, the reward function is handcrafted and must take into account all the possible circumstances that might affect the task to some degree. 

Moreover, the goal, as described by the reward, must be completely specified mathematically.
As the complexity of the task increases, both those aspects become incredibly hard. Therefore, specifying a correct reward function for complex tasks is usually considered an intractable problem \cite{Russell2019}. As a practical solution to this problem, \emph{behavioural cloning} (BC) \cite{TorabiStone2018} and \emph{imitation learning} (IL) \cite{KoberPeters2010,BoutilierPrice2003} approaches are used to tackle it by using demonstrations, or recorded \textit{trajectories}. In these approaches, each trajectory corresponds to a set of observations and actions in the form of a tuple $\mathcal{D}={(o_i, a_i), i=1,\ldots,T}$ recorded from one or more experts, and where $T$ denotes the end of the episode. Those demonstrations are then used as a dataset to train a policy in a supervised learning fashion; at the end of the training procedure, the obtained network represents the actual policy that takes decisions in the RL environment.

Despite showing promise, even these approaches have their limitations \cite{Codevilla2019,KanervistoHautamaki2020,KanervistoPussinen2020}. Among the others, the most important one resides in their inherent dependency from the expert's trajectories assumed optimality. Previous work \cite{KanervistoPussinen2020} shows that BC agents usually struggle in replicating human-level performances, let alone surpassing them. This happens because, despite the agent's ability to generalize over the training examples, a task usually comes with subproblems \cite{Russell2019} whose solutions might not be exemplified in the gathered data. The number of those subproblems exponentially increases with the complexity of the task \cite{Russell2019}, and a dataset that takes into account all those things would be too hard and costly to aggregate and validate.

In 2010, \cite{RossBagnell2010} proposed the \textit{Dataset Aggregation (DAgger) Algorithm} in order to address the problem of providing more representative trajectories in the BC dataset. DAgger addresses this problem by using the BC agent trained on the original dataset as an expert. Then it trains a new BC agent on the new dataset, which will be used to gather even more representative data and so on. The underlying idea of the method is that the problem of not having enough data can be solved by extracting more trajectories from an expert-trained agent. Although the idea seems reasonable, as the newly extracted data would surely follow the expert's data distribution, it is limited by the fact that such an agent is trained on incomplete data. Therefore it is necessarily sub-optimal and progress, if any, might proceed slowly.

One of the evolutions of this algorithm is represented by \textit{Human-Gated Dataset Aggregation Algorithm} \cite{KellyKochenderfer2018} (HG-DAgger). This approach uses the same idea as the previous but includes human feedback in the training loop. HG-DAgger relies on the idea that in order to improve, an agent must be provided with high-quality labels. To gather them, a human expert can take over control during a standard RL game played by the BC agent and provide corrective trajectories. At the end of this correction step, the new demonstrations are added to the original dataset and the agent is fine-tuned on this augmented dataset, similarly to the original DAgger algorithm. We argue that while the method leads to higher quality labels, their effect on the policy highly depends on the BC dataset. This happens because adding these corrections to the dataset means merging their distribution with the one from the original trajectories. As a result, the distribution of the data will change but stay anchored to the original one, unless the number of corrective and baseline actions become comparable. Thus, in cases where the optimal trajectory is very distant from the examples that we have, it won't be possible for the agent to follow it.

In this paper, we propose a novel approach to address e problem and possibly solve it. In our approach, the human expert supervising the algorithm can take over control of a simulation and provide corrective micro-trajectories to the agent similar to the HG-DAgger algorithm. Then, instead of augmenting the dataset, we perform a BC training step on the spot. Such guided training represents an intermediate step between the standard BC training and the testing phases. We assess the validity of our approach on a Minecraft agent and provide both quantitative and qualitative comparison of our agent's performance with respect to pure BC and two DAgger agents trained on the same environment.
\section{Related Work}
Our approach takes the original DAgger algorithm \cite{RossBagnell2010} as the main inspiration. In that work, the authors use the idea of fixing wrong actions of a standard BC agent by re-evaluating its errors through another policy. Although the idea sounds promising, it assumes that the policy used for correction is better than the actual policy at solving the task or a particular instance of it. We argue that such an assumption is strong and perhaps too optimistic, as having such an agent would make the actual agent useless and solve the problem inherently. Moreover, generating an expert agent for each possible problem is to be considered impractical due to the enormous amount of resources it would require for non-trivial tasks \cite{Russell2019}.

Our method uses one of the DAgger variants, HG-DAgger \cite{KellyKochenderfer2018}, as a secondary source of inspiration. This algorithm addresses the problem described above and deals with it by introducing a human expert in the training procedure. The presence of a human expert is used at training time to correct trajectories that show dangerous behaviours. Therefore, the overall quality of the training trajectories increases and the average performance of the agent increases as well. Even though we believe that this algorithm represents a valid solution to the original DAgger problem, we argue that introducing the human expert at training time implies a gigantic work of correction and provides very slow improvement for a newly generated policy. Therefore, a non-trivial task might require a non-feasible amount of work to be done to train a single agent.

Moreover, previous work on {\em inverse reinforcement learning} \cite{ZiebartDey2008} and on RL {\em human feedback} \cite{SinghLevine2019} are taken into account for the development of our method, to provide suggestions on keeping our method fast and reliable. In particular, \cite{ZiebartDey2008} proposes the {\em maximum entropy principle} for exploring the reward space. This principle inspired the idea of detaching the BC dataset from the expert correction so that the data distributions remain separate and provide maximum exploration capabilities to the agent. On a more technical part, \cite{SinghLevine2019} proposes a novel method to make the best use of a low number of trajectories. In our paper, this idea helped us in shaping the training step on the corrective trajectory.

\section{Human-Driven Dynamic Dataset Augmentation}
In this section, we present Human-Driven Dynamic Dataset Augmentation (HDD-DAugment), a method for improving RL agents trained from human demonstrations, that addresses and solves all the problems presented above. Similarly to the HG-DAgger algorithm, the motivation for our approach is to provide high-quality labels to an agent. We believe that these labels can be provided only by an expert, which must be therefore included in the training loop to some extent. However, differently from the baseline approach, we integrate the expert's corrections in the form of one-time micro-trajectories learned on the spot, skipping the dataset augmentation step. 

In the HG-DAgger algorithm described in Algorithm \ref{alg:hg-dagger}, a \textit{novice policy} $\pi_{N_{i}}$ is pre-trained over a dataset $\mathcal{D}_{\mathrm{BC}}$ that is then augmented iteratively based on corrections provided by an expert. The resulting dataset $\mathcal{D}$ is then used to fine-tune the agent by performing additional training epochs in the standard BC fashion. Although the solution is indeed valid and leads to an improvement of the policy, we argue that it might suffer from data-related problems in contexts that deal with high dimensional data such as images. Among those problems, the most prominent is maintaining the data in memory, for which a solution requires performing an infinite number of read/write operations from slow mass storage memories. In those cases, we believe that two main reasons could slow down or even prevent an improvement over a standard BC agent.

First, dealing with high dimensional data drastically increases their variability, and therefore a much more vast set of trajectories is required to allow learning. As a consequence, augmenting the dataset results in a longer training time. Also, memory is affected: a high dimensional dataset might not fit into memory. In those cases, data are stored to mass storage devices, and become an additional cost both in economic and computational terms. Although time can be saved with highly-engineered solutions, memory for massive datasets of this kind is still problematic.

Second, DAgger-like approaches incorporate the corrections into the original dataset $\mathcal{D}_{\mathrm{BC}}$, such that at the end of the procedure $\mathcal{D}_{\mathrm{BC}} \subseteq \mathcal{D}$. This prevents the agent from drifting away from the original solution, as a good portion of the augmented dataset will still represent the original one. We believe that such robustness to change is both an upside and a drawback: whenever the variability of the data is high, preventing the drift might anchor the agent to a solution that is very distant from the optimal one. On the other hand, allowing the agent to detach from it by learning from an independent dataset allows the agent to freely explore new solutions to the problem. Indeed, this comes with a risk of detaching from the initial task, but the complete control that the expert exerts over the agent helps in preventing this.

By skipping the standard aggregation step, our approach avoids the data-related problems and can be used to train a network on high dimensional datasets: HDD-DAugment (Algorithm \ref{alg:hdd-daugment}) does not make use of the initial BC dataset, thus making the intermediate training steps much faster than the baseline approaches. Our method processes an expert's corrective trajectory as an independent mini-batch of $(o_t, a_t)$ tuples that are stored only temporarily in RAM and are deleted as soon as the training step ends. As a consequence of this, we do not need to perform onerous read and write actions, nor occupy memory with corrective data or need a powerful machine to overcome those limitations. At the same time, using the corrective trajectory independently from the baseline dataset makes it independent from the underlying data distribution. Therefore, the policy can explore a wider range of solutions for the problem.

While this might come with problems if the number of corrections is too high, a single correction does represent a small step in an arbitrary direction and is equivalent to the idea of random noise added to the policy in evolutionary algorithms \cite{Grefenstette2011}. Thus, small changes to the policy weights can be allied with no harm. Moreover, since each correction is provided singularly by an expert, a careful design of corrective trajectories and limiting their number can easily avoid this negative phenomenon.

\begin{algorithm}[tb]
\caption{HG-DAgger}
\label{alg:hg-dagger}
\textbf{Input}: $\pi_{H}, \pi_{N_{i}}, \mathcal{D}_{\mathrm{BC}}$\\
\textbf{Output}: $\pi_{N_{K+1}}, \tau$
\begin{algorithmic}[1] 
\STATE $\mathcal{D} \leftarrow \mathcal{D}_{\mathrm{BC}}$
\STATE $\mathcal{I} \leftarrow []$
\FOR{epoch $i = 1 : K$}
\FOR{rollout $j = 1 : M$}
\FOR{timestep $t \in T$ of rollout $j$}
\IF{expert has control}
\STATE Record expert labels into $D_{j}$
\ENDIF
\IF{expert is taking control}
\STATE record doubt into $\mathcal{I}_{j}$
\ENDIF
\ENDFOR
\STATE $\mathcal{D} \leftarrow \mathcal{D} \cup \mathcal{D}_{j}$
\STATE $\mathcal{I} \leftarrow \mathcal{I} \cup \mathcal{I}_{j}$
\ENDFOR
\STATE train $\pi_{N_{i+1}} on \mathcal{D}$
\ENDFOR
\STATE $\tau \leftarrow f(\mathcal{I})$
\STATE \textbf{return} $\pi_{N_{K+1}}, \tau$
\end{algorithmic}
\end{algorithm}

\begin{algorithm}[tb]
\caption{HDD-DAugment}
\label{alg:hdd-daugment}
\textbf{Input}: $\pi_{\mathrm{exp}}, \pi$\\
\textbf{Output}: $\pi$
\begin{algorithmic}[1] 
\STATE train $\pi$ on $\mathcal{D}_{\mathrm{BC}}$ with regular BC
\FOR{game $i = 1 : K$}
\WHILE{not ending condition}
\FOR{timestep $t \in T$ of rollout $j$}
\STATE perform RL step
\IF{expert takes control}
\STATE $\mathcal{D}_{\mathrm{exp}} \leftarrow []$
\WHILE{expert gives control back}
\STATE $\mathcal{D}_{\mathrm{exp}} \leftarrow \mathcal{D}_{\mathrm{exp}} \cup \{(o_{exp}, a_{\mathrm{exp}})\}$
\ENDWHILE
\STATE train $\pi$ on $\mathcal{D}_{\mathrm{exp}}$
\ENDIF
\ENDFOR
\ENDWHILE
\ENDFOR
\STATE \textbf{return} $\pi$
\end{algorithmic}
\end{algorithm}

\section{Experiments}
We have performed our experiments on a Minecraft agent, using the MineRL \cite{GussSalakhutdinov2019} dataset. In particular, our agents have been trained to solve the FindCave task proposed in the NeurIPS BASALT challenge \cite{ShahDragan2021} 2021. To complete this task, an agent must spot a cave and take a picture of it by throwing a snowball while being in front of it. The experiments are focused on two sub-problems that we have discovered during the competition. In particular, we found out that standard BC agents were struggling to jump on ramps of blocks and handle the camera controls. To address them, we propose a comparison between four agents. As the first step, each policy has been trained with standard BC until convergence. From our experiments, this happened after around $150$ epochs. Then, three of them were fine-tuned with DAgger, HG-DAgger and HDD-DAugment respectively, while the fourth one was left unchanged.

For each fine-tuned agent we have gathered a similar number of additional data. In particular, the DAgger and the HG-DAgger agents have been trained for $15$ iterations, each one consisting of one game played by the respective policy and $5$ additional training epochs. In both cases, this led to an increment of about $10\%$ in the size of our dataset. To reach a similar number of additional data, we have played $58$ games with the agent trained with HDD-DAugment. After the training, we have collected $20$ games played by each agent, with a maximum duration of $3$ minutes each. Since comparing trajectories qualitatively is not easy, we have decided to use a fixed seed on our environment, to set the same spawning point for every agent. We have assessed the agents' performance according to four different metrics.

\begin{figure*}[t]
\centering
\includegraphics[width=0.9\textwidth]{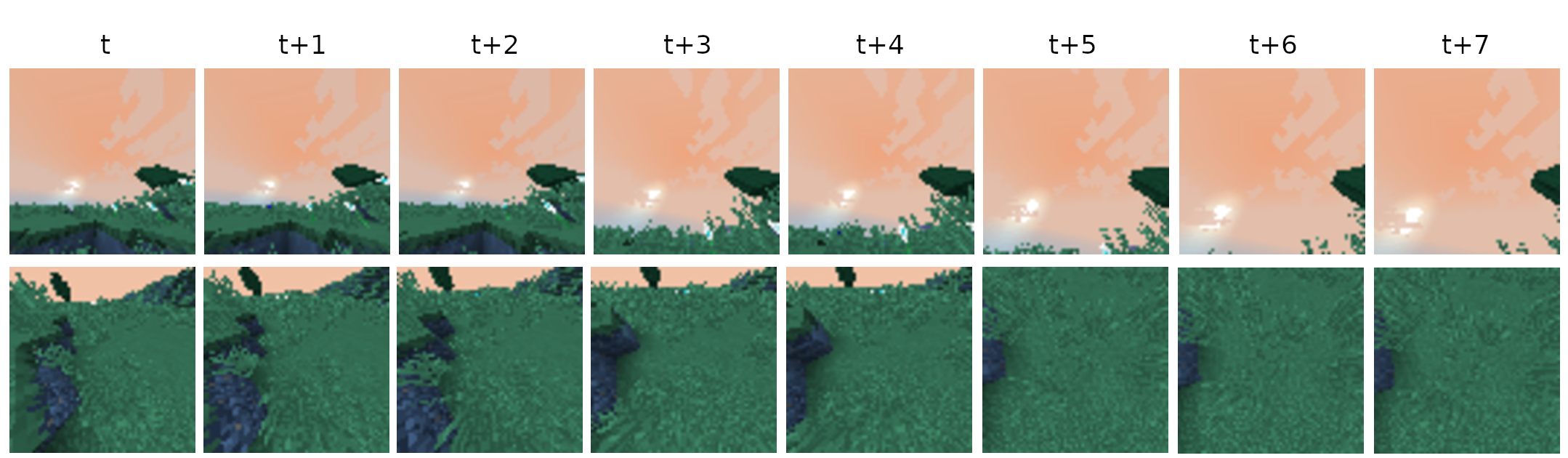}
\caption{Examples of trajectories that lead to blinded situations. \textit{Above:} the agent stares at the sky and loses almost every spatial reference of its immediate surroundings; \textit{Below:} the agent lowers its gaze at every timestep, until it sees nothing but grass.}
\label{fig:blind}
\end{figure*}

\paragraph{Number and severity of collisions.} 
We define a {\em collision} as a situation in which an agent gets stuck on a wall or a block. We evaluate the collisions by dividing them into severity classes, according to their duration: the longer the collision, the higher its severity. In our experiments, we have measured collisions that have lasted more than $5, 10, 20, 30, 45, 60, 90, 120$ and $150$ seconds. An occurrence in the $5$s collision means that in one case an agent has been stuck for a period of time $t$ s.t. 5 $\leq t < 10$, and similarly for all the other values. In this experimental setup, a single game can last at most $180$ seconds. We compare the agents by assessing the total time spent colliding in each game and plotting their values over the total number of games. In this way, we aim to penalize both a single collision that lasts for a very long time and many collisions that are resolved quickly.
\paragraph{Number and severity of blinded situations.} A blinded situation is a particular state in which the agent repeatedly moves the camera in the same direction along the Y-axis. Examples of this situation are represented in Figure \ref{fig:blind}. In these situations, the agent ends up looking at the sky or staring at the ground and loses the vast majority of information about its surroundings. The increment of the camera in each direction is constant and in our case has been set to $5$ degrees per action. To evaluate a blinded situation we set the initial camera position on the vertical axis to $0$ degrees and keep track of the evolution of the offset throughout the game. If the agent overcomes a critical threshold, then it reaches the "blinded state". In our experiments, the values of the threshold have been changed incrementally from $5$ to $90$ degrees with a step of $5$ degrees. Similarly to the number of collisions, we sum all the instances related to a game and plot the sequence with respect to the number of games to assess the agents' tolerance to this problem.
\paragraph{Similarity with expert trajectories.} Provided that the human player is the expert, we know the shape of the optimal solution to a particular situation. For our experiments we have decided to rely on two situations: on the first, the agent is required to step over a block, while the second one assesses an agent that is stuck against a wall. By analyzing the gathered data, we have found those to be the more reliable measures as both of them had an adequate number of instances. For each of them, we extract $20$ instances of the problem from the test games and compute an average trajectory from the one performed by the agents. Then, we compare them with the sequence of actions provided by the expert to overcome the problem. To ensure fairness, we align the first meaningful action to solve the situation and compute the average trajectory accordingly.
\paragraph{Qualitative evaluation.} Since this work has been inspired by the BASALT competition \cite{ShahDragan2021}, and since human likeliness is the focal point of the competition, we also provide a link to a YouTube video that shows the qualitative performance related to each agent. In it, each agent is shown while performing a limited set of actions. While this does not represent a reliable assessment of performances, we believe it is still valid as a perceptual measure of the quality of an agent. The link to the video is posted at \cite{youtube}.
\section{Results}
\begin{figure}[t]
\centering
\includegraphics[width=0.9\columnwidth]{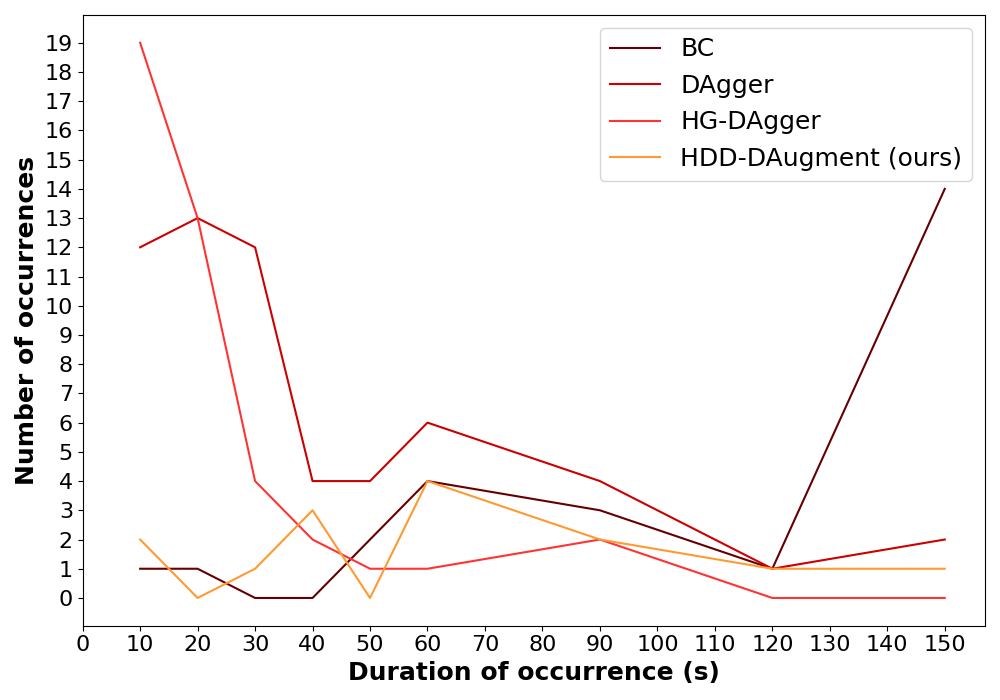}
\caption{The graph shows the number of occurrences ($y$-axis) per each measured duration ($x$-axis) of the four approaches we compare. In our study, an occurrence is defined as a general instance of problem that the agent faces (e.g. getting stuck, looping with camera movement and so on). The BC agent gets stuck for very long amounts of time in 14 games out of 20. Both the DAgger and HG-DAgger policies show a high number of quick collisions, that rapidly decrease as the duration increases. The HDD-DAugment agent shows a low number of collisions, regardless of their length.}
\label{fig:occs}
\end{figure}
\begin{figure}[t]
\centering
\includegraphics[width=0.9\columnwidth]{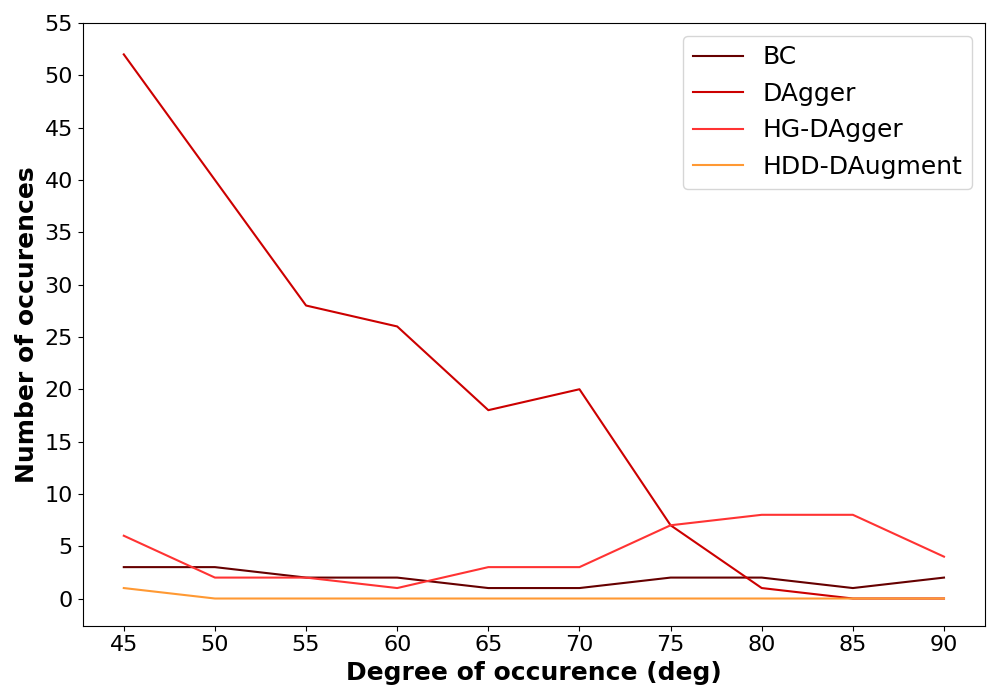}
\caption{The number of blinded situations ($y$-axis) for each agent, computed with respect to different angle thresholds ($x$-axis). The DAgger agent shows a linearly decreasing trend. The HG-Dagger policy has some occurrences ($<10$) for very high thresholds.}
\label{fig:deg_occs}
\end{figure}
\paragraph{Number of collisions.} 
Figure \ref{fig:occs} shows the number of collisions per each measured duration. The HDD-DAugment agent handles collisions in the best way possible. While the number of quick collisions is comparable to the baseline BC agent, it is noticeable that the trend of our agent does not vary as the duration of the collision increases. On the other side, the trend of the BC agent skyrockets, meaning that it often gets stuck indefinitely and wastes the majority of its exploration time. Moreover, our agent gets mostly stuck in situations that include a pond or a swap, from which is particularly hard to get out. Our approach also outperforms both the DAgger and the HG-DAgger policies, which suffer from a high number of non-severe collisions.

\begin{figure}[t]
\centering
\includegraphics[width=0.9\columnwidth]{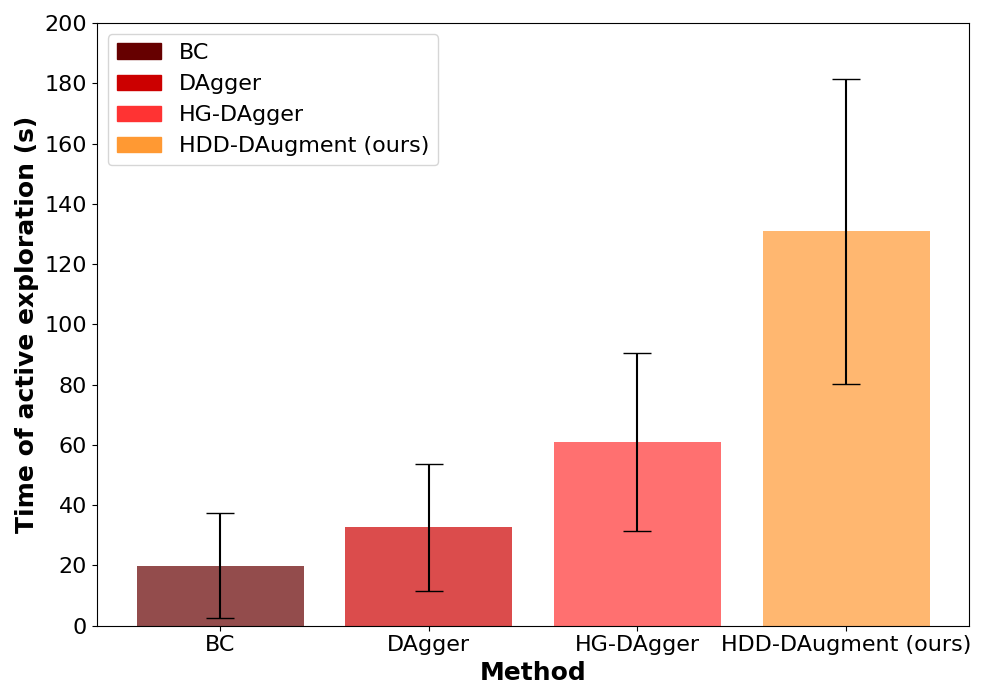}
\caption{Number of frames ($y$-axis) that have been used for active exploration in each game ($x$-axis). The BC agent spends most of its time being stuck, along with the DAgger policy. Having a lot of quick collisions, also the HG-DAgger is penalized with this measure. Our HDD-DAugment policy is the only one to reach the maximum uptime in some games and rarely drops under $60\%$ of uptime.}
\label{fig:uptime}
\end{figure}

Moreover, Figure \ref{fig:uptime} shows that the improvement of our agent with respect to the other candidates is quite noticeable. In particular, our agent is the only one that never gets stuck in some of the games. Also, even in the worst-performing game, HDD-DAugment performs similarly to the best HG-DAgger. The image shows clearly that on average our method outperforms all the others in terms of active time of exploration.
\begin{figure*}[t]
\centering
\begin{tabular}{ccc}
\includegraphics[width=0.47\textwidth]{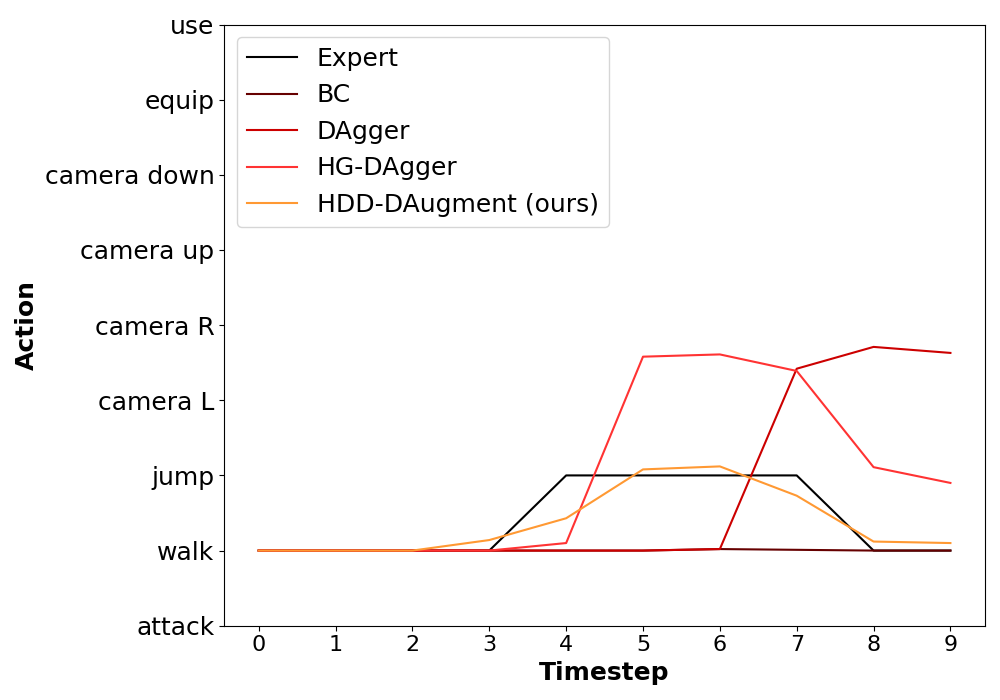} &
\includegraphics[width=0.47\textwidth]{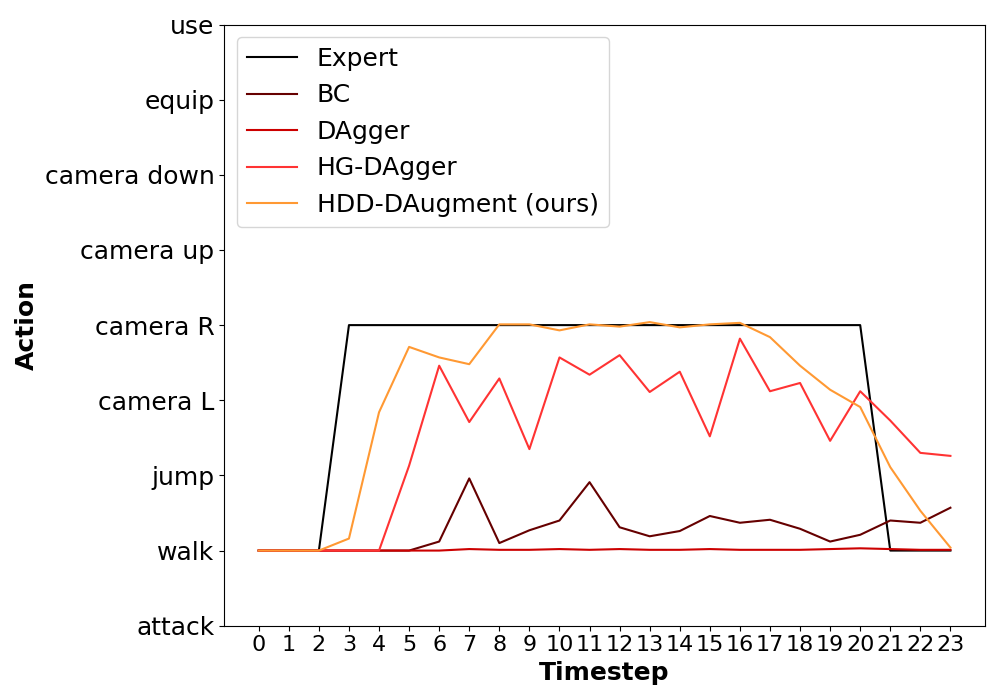} \\
\textbf{(a)}  & \textbf{(b)} \\[6pt]
\end{tabular}
\caption{
\textbf{(a)} The average behaviour varies enormously between the agents even when they encountered a single step block. The image shows that while the BC policy continues to walk forward indefinitely, the DAgger and the HG-DAgger agents both try to tilt their cameras to get rid of the obstacle. The HDD-DAugment, instead, replicates almost perfectly the expert's policy.
\textbf{(b)} When an agent faces a wall, the expert turns back and continues walking. The graph shows that the BC and DAgger agents are unable to face this situation. On the other hand, the HG-DAgger policy tilts its camera left and right most of the time. On the contrary, our agent turns the camera with a reasonable degree of certainty for several consecutive timesteps.}
\label{fig:evaluations}
\end{figure*}
\paragraph{Number of blinded situations.} From Figure \ref{fig:deg_occs} we can conclude that the camera management represents a problem only for the BC agent, while all the other approaches successfully solve it. Even so, the HDD-DAugment performs the better on the test games, showing practically no instance of the problem. It is also noticeable how to HG-DAgger almost always recovers for slightly failures of this kind, while if it goes above a higher threshold of tilting, the chances of repairing this problem decrease. This might indicate that the HG-DAgger algorithm solves the problem only if it can look at its surroundings, while a severe blinded situation still represents an impossible problem. Moreover, the low number of occurrences in the HDD-DAugment agent indicates that drifting from the baseline policy is not likely to happen, as we expected.
\paragraph{Similarity with expert trajectories.} In Figures \ref{fig:evaluations}a and \ref{fig:evaluations}b it is visible that the HDD-DAugment agent outperforms all the others both in jumping a step and overcoming a rock wall. It is particularly of interest the fact, also shown in the related video, that our agent mimics the expert almost perfectly: when it faces a wall, it turns by almost $180$ degrees and keeps walking as nothing happened. This is a clear example of a situation in which drifting from the baseline policy comes with major advantages and helps our agent solve situations that none of the others could address easily. Moreover, this example shows that our approach can follow the expert lead closely while not forgetting the knowledge derived from the baseline training.
\paragraph{Qualitative evaluation.} From the video, it is evident that our HDD-DAugment agent is more human-like than all the other agents. It is surprising, perhaps, to see that the HG-DAgger is quite capable of exploring, although it shows its robotic nature in actions like jumping over a block or climbing a hill. Also, we can see that both the baseline BC and the DAgger agents struggle to explore the environment and get stuck most of the time, even though sometimes they can solve situations like stepping on a block.

The video also shows situations for which, unfortunately, the number of examples was so low that building a statistic on it would have led to vague and non-meaningful results. Anyway, the differences between our agent and the others are clear. It is particularly interesting to see how our agent manages to get out of a pond: while every other agent usually has to be lucky to find the correct path, our agent spots the correct path and follows it. Finally, it's visible that our agent from time to time tilts its camera left and right while exploring. We have decided to include this in the video to show the possible negative effects of drifting from the baseline policy. In our experience, those problems come from non-coherent solutions provided by a human expert. Therefore, it is possible to avoid them easily with a careful design of the corrective trajectories.
\section{Conclusions}
In this paper, we propose {\em Human-Driven Dynamic Dataset Augmentation} (HDD-DAugment), a novel approach that incorporates human feedback to address and tackle problems of previous algorithms for improving the performances of a {\em behavioural cloning} (BC) agent. Moreover, we apply our approach to a standard BC agent and compare it with the baseline BC policy itself and two other similar approaches, DAgger and HG-DAgger. We evaluate the performances of each approach with both qualitative and quantitative measures.

Our experimental work shows that while our agent is far from being perfect, it either matches or improves the other approaches in all the instances of problems that we have assessed. In some cases, our agent was the only one that could provide a solution. The obtained results prove that even though HDD-DAugment is a more risky approach than the others, it comes with great upsides and careful planning of the human corrections greatly reduces those risks. Moreover, our algorithm is invariant to problems related to the storage of data and the speed of training and provides a fast and light way to train an agent even with extensive, high dimensional datasets.
\bibliography{main.bib}
\end{document}